\pdfoutput=1

\documentclass[11pt]{article}

\usepackage[final]{acl}
\usepackage{float}
\usepackage{times}
\usepackage{latexsym}
\usepackage{booktabs} 
\usepackage{multirow}
\usepackage[T1]{fontenc}

\usepackage[utf8]{inputenc}
\usepackage{cleveref}
\crefformat{section}{\S#2#1#3}

\usepackage{microtype}

\usepackage{graphicx}

\newcommand{\datasetname}{\texttt{ConECT}}
\newcommand{\allegro}{\texttt{allegro.pl}}
\newcommand{\mall}{\texttt{mall.cz}}

\title{\datasetname~Dataset: Overcoming Data Scarcity in Context-Aware E-Commerce MT}

\author{
 \textbf{Mikołaj Pokrywka\textsuperscript{1,2}},
 \textbf{Wojciech Kusa\textsuperscript{1,3}},
 \textbf{Mieszko Rutkowski\textsuperscript{1}},
 \textbf{Mikołaj Koszowski\textsuperscript{1}}
\\
 \textsuperscript{1}Machine Learning Research Allegro.com,
 \textsuperscript{2}Laniqo.com,
 \textsuperscript{3}NASK
\\
 \small{
   \textbf{Correspondence:} \href{mailto:@domain}{\{name\}.\{surname\}@allegro.com}
 }
}

\begin{document}
\maketitle
\begin{abstract}
Neural Machine Translation (NMT) has improved translation by using Transformer-based models, but it still struggles with word ambiguity and context. 
This problem is especially important in domain-specific applications, which often have problems with unclear sentences or poor data quality. 
Our research explores how adding information to models can improve translations in the context of e-commerce data.
To this end we create \datasetname --  a new Czech-to-Polish e-commerce product translation dataset coupled with images and product metadata consisting of 11,400 sentence pairs.
We then investigate and compare different methods that are applicable to context-aware translation.
We test a vision-language model (VLM), finding that visual context aids translation quality. 
Additionally, we explore the incorporation of contextual information into text-to-text models, such as the product's category path or image descriptions. 
The results of our study demonstrate that the incorporation of contextual information 
leads to an improvement in the quality of machine translation.
We make the new dataset publicly available.\footnote{\url{https://huggingface.co/datasets/allegro/ConECT}}

\end{abstract}

\section{Introduction}

Neural Machine Translation (NMT) has significantly advanced the field of machine translation by leveraging Transformer-based models \cite{bahdanau2014neural,vaswani2017attention}. These models have been critical in enhancing translation quality, particularly by incorporating mechanisms such as cross-attention to achieve better semantic understanding. However, despite these improvements, sentence-level translation in NMT often struggles with issues such as contextual disambiguation~\cite{rios-gonzales-etal-2017-improving}. For example, the word \textit{``pen''} can refer to a writing instrument or an enclosure for animals, depending on the context. These ambiguities present a significant challenge in achieving accurate translations based solely on the context of single sentences.

\begin{figure}[t]
    \centering
    \includegraphics[width=\linewidth]{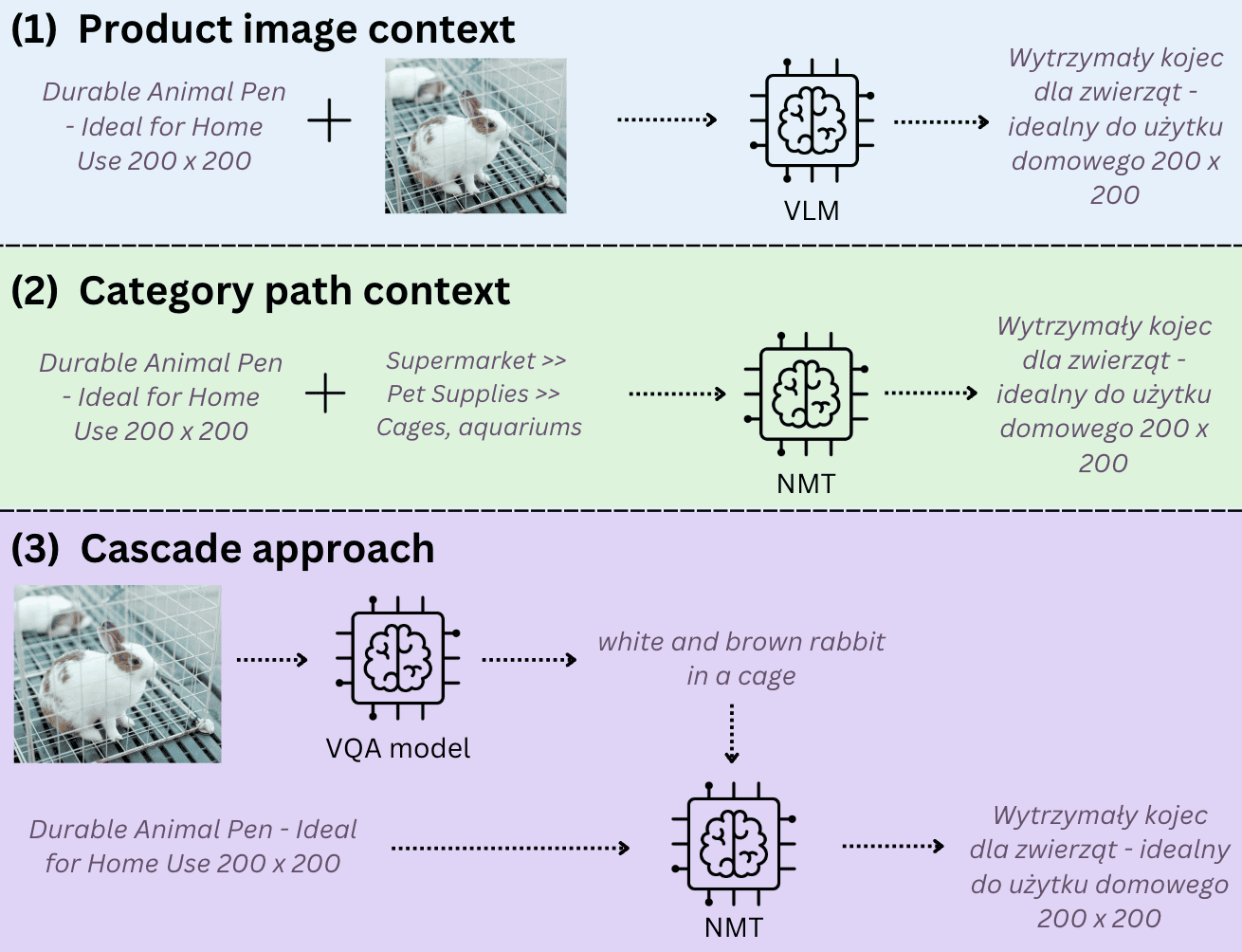}
    \caption{We evaluate three methods for contextualisation in e-commerce MT: (1) combining images with text in VLM, (2) appending category path context in NMT, and (3) a cascade approach consisting of a vision Q\&A and a text-to-text NMT.}
    \label{fig:abstract}
\end{figure}

In an attempt to address these limitations, Multimodal Machine Translation (MMT) has emerged as a promising paradigm~\cite{specia-etal-2016-shared,shen2024survey}. MMT integrates visual information alongside textual data to provide additional context, thereby enhancing translation quality. Studies have shown that leveraging visual cues can significantly improve the disambiguation of lexical items and contribute to more accurate translations \cite{yao-wan-2020-multimodal,wang2021efficient}. For instance, visual context can help resolve ambiguities in sentences where the textual information alone is insufficient \cite{liu2021gumbel}. 
However, high-quality machine translation training data with aligned images is available only for a fraction of the parallel corpora.

In this work, we explore the effectiveness of integrating context information into a domain-specific translation task. We create a new testset for Czech-to-Polish translation in e-commerce, and test three approaches for context-aware MT (Figure~\ref{fig:abstract}). We fine-tune and evaluate a VLM, as well as traditional NMT models for which we integrate contextual information as special instruction tokens.
Our findings demonstrate that visual context can enhance translation quality in domain-specific scenarios, serving as a valuable additional feature. 

Our contributions are as follows:
\begin{itemize}
        \item We release a new multimodal dataset for e-commerce MT for the under-researched cs-pl language pair;
        \item We fine-tune and evaluate a VLM on the domain-specific MT task;
        \item We propose a method for encoding category paths and image descriptions in small NMT models.
\end{itemize}

We start by describing related work (\cref{sec:related}), then introduce the \datasetname~dataset (\cref{sec:dataset}), explain our experimental setup (\cref{sec:setup}), and finally discuss the results (\cref{sec:results}).

\section{Related work} 
\label{sec:related}

In this section we discuss multimodal and e-commerce product-oriented MT. 

\subsection{Multimodal Machine Translation}
Multimodal Machine Translation (MMT), which integrates visual context into machine translation, has garnered significant attention in recent years~\cite{elliott-etal-2016-multi30k,shen2024survey}. Research has shown that visuals can effectively bridge linguistic gaps between languages \cite{chen2019words,sigurdsson2020visual,su2019unsupervised}. Early efforts by \citet{calixto-liu-2017-incorporating} incorporated global image features into the encoder or decoder of NMT models. Subsequent studies have explored the use of more granular image contexts, such as spatial image regions with attention mechanisms~\cite{calixto-etal-2017-doubly,caglayan2016multimodal}. 

\citet{yang2020visual} introduced a method for joint training of source-to-target and target-to-source models to promote visual agreement. \citet{yin-etal-2020-novel} built a multi-modal graph linking image objects and source words, leveraging an external visual grounding model for alignment.

\subsection{Benchmarks and Shared Tasks in MMT}

Between 2016 and 2018, WMT organised a shared task on MMT~\cite{specia-etal-2016-shared,elliott-etal-2017-findings,barrault-etal-2018-findings}.
The organisers primarily used the Multi30K dataset~\cite{elliott-etal-2016-multi30k} and metrics like BLEU, METEOR, and TER. 
In the first edition of the shared task, \citet{specia-etal-2016-shared} found that neural MMT models initially underperformed compared to text-only SMT models. However, \citet{elliott-etal-2017-findings} and \citet{barrault-etal-2018-findings} expanded languages and test datasets, noting performance improvements in MMT models, especially when incorporating external resources.

\citet{elliott-2018-adversarial} introduced an adversarial evaluation method to assess whether multimodal translation systems effectively utilize visual context. This method evaluates the performance difference of a system when provided with either a congruent or an incongruent image as additional context.

\citet{futeral-etal-2023-tackling} introduced the \textit{CoMMuTE} (Contrastive Multilingual Multimodal Translation Evaluation) dataset to evaluate multimodal machine translation systems with a focus on resolving ambiguity using images. Their approach, which includes neural adapters and guided self-attention, showed significant improvement over text-only models, particularly in English--French, English--German, and English--Czech translations.
\citet{futeral2024towards} extended the setup with \textit{ZeroMMT}, a technique for zero-shot multimodal machine translation that does not rely on fully supervised data. This method, which uses visually conditioned masked language modelling and Kullback-Leibler divergence training, demonstrated near state-of-the-art performance and was extended to Arabic, Russian, and Chinese.

\subsection{E-commerce product-oriented MT}
An e-commerce product-oriented machine translation task uses product images and product metadata as inputs.
\citet{calixto-etal-2017-using} pioneered this task with a bilingual product description dataset, evaluating models such as phrase-based statistical MT (PBSMT), text-only MT, and MMT. Their results highlighted PBSMT's superior performance, with MMT models enhancing translation when re-ranking PBSMT outputs.
\citet{calixto-etal-2017-human} highlight the potential impact of multi-modal NMT in the context of e-commerce product listings. With only a limited amount of multimodal and multilingual training data available, both text-only and multi-modal NMT models failed to outperform a productive SMT system.

\citet{song2021product} introduced a large-scale dataset along with a unified pre-training and fine-tuning framework, proposing pre-training tasks for aligning bilingual texts and product images. 
These tasks include masked word reconstruction with bilingual and image context, semantic matching between text and image, and masking of source words conveying product attributes. This framework contributed to more robust translation models.

\subsection{Visual information integration in LLMs}

The field of NLP has evolved with the introduction of large language models (LLMs) \cite{ouyang2022training,achiam2023gpt,touvron2023llama}. Since traditional text-only MT has advanced to LLM-based methods~\cite{xu2023paradigm}, leveraging LLMs for MMT is a promising direction. Current multi-modal LLMs, whether using linear models \cite{liu2024visual} or Query Transformers \cite{li2023blip}, often suffer from visual information loss. Enhancing alignment between textual and visual modalities and adaptively extracting relevant visual information are critical for optimizing LLM performance in MMT tasks. 

\section{\datasetname~dataset}
\label{sec:dataset}

The \datasetname~dataset (\textbf{Con}textual \textbf{E}-\textbf{C}ommerce \textbf{T}ranslation) is designed to support research on context-aware MT in the e-commerce domain.
To create this dataset, we extracted 11,000 sentences in Polish from the \allegro~e-commerce platform.
Next, we aligned a primary product image and category path. %
The sentences were then manually translated into Czech by professional translators. 
Each translation was reviewed to ensure accuracy and contextual relevance. 
A detailed breakdown of the dataset statistics is provided in Table~\ref{tab:dataset_statistics}.

The \datasetname~dataset is divided into various content types to provide comprehensive coverage of e-commerce translation contexts, as summarized below.

\paragraph{Product names}
Product names are typically short phrases that precisely identify a product. They often contain specific terminology, brand names, and product specifications.

\paragraph{Product descriptions}
Product descriptions are longer texts that provide detailed information about a product, including its features, specifications, usage instructions, and benefits. These texts can be more descriptive and less structured than product names.

\paragraph{Offer titles}
Offer titles are concise and attractive phrases crafted by marketers to engage potential buyers, often including promotional language, discounts, or special offers. This category can be challenging due to the need to maintain both the persuasive tone and the specific promotional content during the translation process.

\begin{table}[]
\resizebox{\columnwidth}{!}{%
    \centering
\begin{tabular}{llrrr|rrr}
\toprule
& Content & & & \#category & \multicolumn{3}{c}{CS} \\
Split & type & \#Sent. & \#Img. & paths & \#Tokens & Len. & Vocab  \\
\midrule
\multirow{4}{*}{Test} & Offer titles & 1,924 & 1,920 & 840 & 13,898, & 7.2 & 6,206\\
 & Prod. desc. & 3,680 & 2,905 & 1,090 & 38,885, & 10.6 & 13,286 \\
 & Prod. names & 4,691 & 4,659 & 1,361 & 34,886, & 7.4 & 10,422 \\
 & ALL & 10,295 & 6,146 & 1,542 & 87,669, & 8.5 & 22,121 \\ \midrule \midrule
\multirow{4}{*}{Valid} & Offer titles & 203 & 203 & 165 & 1,449 & 7.1 & 1,080  \\
 & Prod. desc. & 403 & 389 & 285 & 4,118 & 10.2 & 2,472  \\
 & Prod names & 505 & 505 & 360 & 3,757 & 7.4 & 2,116  \\
 & ALL & 1,111 & 1,042 & 596 & 9,324 & 8.4 & 4,822 \\
\bottomrule
\end{tabular}
}
    \caption{\datasetname~dataset statistics. \textit{Len.} denotes average length in words. Polish sentences have similar statistics.}
    \label{tab:dataset_statistics}
\end{table}

\section{Experimental setup}
\label{sec:setup}

In this section we describe the training and evaluation data, the models used, and the evaluation procedure.

\subsection{Training data}

To create a parallel e-commerce dataset, we aligned Polish and Czech product names and descriptions for corresponding products that were listed on both e-commerce platforms \allegro~ and \mall~. Merchants manually translated original Czech product names and descriptions into Polish. To create sentence-level pairs of multi-sentence descriptions, we split them into sentences and aligned them using a language-agnostic BERT sentence embedding model~\cite{feng2022languageagnosticbertsentenceembedding}. We compared every sentence in one description with every sentence in the corresponding description in the other language to align the sentence pairs. That procedure resulted in a dataset containing 230,000 parallel sentences, each paired with product category paths and one of 38,000 unique images.

\paragraph{Data with image context}
\label{data_vlm}
For experiments involving translation with image context, we additionally collected 440,000 Polish product names paired with 430,000 unique images from the \allegro~e-commerce platform. These product names were back-translated into Czech, creating a synthetic dataset of product names with image context.
The images were in \textsc{JPEG} format with varying sizes and were resized to 224x224 pixels.

\paragraph{Text-to-text models}
For the baseline text-to-text models, we used 53 million sentence pairs, primarily drawn from the OPUS corpora~\cite{tiedemann-nygaard-2004-opus} and internal e-commerce domain data.

For fine-tuning with category paths as context, we extended the 230,000 parallel sentences from the original dataset with 7 million back-translated product names and 7 million back-translated product description sentences, paired with their respective category paths. Additionally, we extended the fine-tuning dataset by incorporating 7 million parallel sentences without category paths from the baseline model’s training set. The category paths were represented as text, listing the hierarchical subcategories for each product (e.g., "\textit{Sports} >> \textit{Bicycles} >> \textit{Tires}").

For experiments involving image descriptions, we generated image descriptions in Czech using the~\texttt{paligemma-3b-mix-224}\footnote{\url{https://huggingface.co/google/paligemma-3b-mix-224}}~\cite{beyer2024paligemmaversatile3bvlm} model on all of the previously mentioned data with image context. Additionally, for fine-tuning, we included 700,000 sentences without image descriptions, extracted from the baseline model's training set.

\begin{table*}[!htp]
\resizebox{\textwidth}{!}{%
    \centering
\begin{tabular}{lll|rr|rr|rr|rr}
\toprule
\multirow{2}{*}{Model}                                                                 & \multirow{2}{*}{Train} & \multirow{2}{*}{Inference} & \multicolumn{2}{c|}{Product names}                    & \multicolumn{2}{c|}{Offer titles}                     & \multicolumn{2}{c|}{Product desc}                     & \multicolumn{2}{c}{All sets}                         \\ \cline{4-11} 
                                                                                       &                        &                            & \multicolumn{1}{c}{chrF} & \multicolumn{1}{c|}{COMET} & \multicolumn{1}{c}{chrF} & \multicolumn{1}{c|}{COMET} & \multicolumn{1}{c}{chrF} & \multicolumn{1}{c|}{COMET} & \multicolumn{1}{c}{chrF} & \multicolumn{1}{c}{COMET} \\ \midrule
NLLB-600M                                                                              & --                     & --                         & 48.46                    & 0.7214                     & 38.01                    & 0.6537                     & 48.50                     & 0.7774                     & 46.85                    & 0.7288                    \\ \midrule
\multirow{4}{*}{PaliGemma-3b}                                                          & real img              & real img                  & \textbf{83.48}           & \textbf{0.9310}             & \textbf{79.41}           & \textbf{0.9083}            & \textbf{61.92}           & 0.8987                     & \textbf{72.31}           & \textbf{0.9152}           \\
                                                                                       & real img              & black img                 & 81.36                    & 0.9224                     & 77.10                     & 0.8972                     & 61.75                    & \textbf{0.8994}            & 71.12                    & 0.9095                    \\
                                                                                       & black img             & real img                  & 82.69                    & 0.9275                     & 77.86                    & 0.9009                     & 60.57                    & 0.8891                     & 71.15                    & 0.9088                    \\
                                                                                       & black img             & black img                 & 82.49                    & 0.9268                     & 77.97                    & 0.9009                     & 60.87                    & 0.8908                     & 71.24                    & 0.9091                    \\ \midrule
Baseline                                                                               & --                     & --                         & 84.83                    & 0.9326                     & 83.73                    & 0.9227                     & 70.76                    & 0.9335                     & 77.74                    & 0.9311                    \\ \midrule
\multirow{2}{*}{\begin{tabular}[c]{@{}l@{}}Category paths\\ experiements\end{tabular}} & \multicolumn{2}{l|}{no category context}            & 85.27                    & 0.9372                     & 83.66                    & 0.9242                     & \textbf{72.78}           & 0.9389                     & \textbf{78.87}           & 0.9354                    \\
                                                                                       & \multicolumn{2}{l|}{category context}               & \textbf{85.51}           & \textbf{0.9385}            & \textbf{83.73}           & \textbf{0.9248}            & 71.95                    & \textbf{0.9393}            & 78.56                    & \textbf{0.9362}           \\ \midrule
\multirow{2}{*}{\begin{tabular}[c]{@{}l@{}}Image desc.\\ experiments\end{tabular}}      & \multicolumn{2}{l|}{no description context}         & \textbf{85.10}            & \textbf{0.9367}            & \textbf{83.99}           & \textbf{0.9246}            & \textbf{70.81}           & \textbf{0.9358}            & \textbf{77.90}            & \textbf{0.9341}           \\
                                                                                       & \multicolumn{2}{l|}{description context}            & 83.25                    & 0.8673                     & 82.63                    & 0.8974                     & 48.26                    & 0.7243                     & 65.97                    & 0.8219                    \\ \bottomrule
\end{tabular}
}
\caption{Comparison of the results on the \datasetname~ test set shows that the VLM model with image context and the NMT model with category paths achieved improved performance due to the added context. However, experiments with synthetic image descriptions led to a decrease in metrics.}
\label{table_results}
\end{table*}

\subsection{Models}
\paragraph{Models with image context}
Experiments involving translation with image context were conducted on the \texttt{paligemma-3b-pt-224}\footnote{\url{https://huggingface.co/google/paligemma-3b-pt-224}} model. Our primary goal was to evaluate the influence of images on translation. 
To achieve this, we fine-tuned the models for the translation task with two types of image data: (1) original corresponding product images, and (2) a black image unrelated to the text input. Further details of the experimental setup are given in Appendix~\ref{app:vlm_ft}.

\paragraph{Text-to-text baseline model}
We developed a sentence-level text-to-text baseline model using the Transformer (big) architecture~\cite{NIPS2017_3f5ee243}, trained with the Marian framework~\cite{mariannmt}. Details of the experimental setup are given in Appendix~\ref{marian_baseline}.

\paragraph{Models with category context}
To ensure a fair comparison, we implemented two fine-tuning approaches on the baseline model: one that incorporates category context and one that does not.
The category context was integrated by adding a prefix containing the product's category path in Polish to the source sentences. The category path was enclosed with special tokens <SC> and <EC> to separate it clearly from the source sentence. The model without category context was trained using the same configuration and data setup, but without the category path prefixes. Details of the experimental setup are given in Appendix~\ref{app:category_context}.

\paragraph{Models with image descriptions}
Following a similar approach as in the category context experiments, we fine-tuned the baseline model with and without prefixes containing image descriptions. The image descriptions were added as prefixes wrapped with <SD> and <ED> tokens. This model was trained exclusively on data with image context converted into image descriptions and did not include any data from category context experiments. Details of the experimental setup are given in Appendix~\ref{app:gen_desc}.

\paragraph{NLLB-baseline}
For comparison, we report our results alongside the NLLB-200-Distilled-600M model.\footnote{\url{https://huggingface.co/facebook/nllb-200-distilled-600M}}

\paragraph{Evaluation metrics}
We use sacreBLEU~\cite{post-2018-call} to calculate the chrF\footnote{\scriptsize chrF signature: nrefs:1$\vert$case:mixed$\vert$eff:yes$\vert$nc:6$\vert$nw:0$\vert$space:no$\vert$version:2.3.1}~\cite{popovic-2015-chrf} score, and the Unbabel/wmt22-comet-da\footnote{\scriptsize Python3.9.19|Comet2.2.2|fp32|Unbabel/wmt22-comet-da}~\cite{rei-etal-2022-comet} model to calculate the COMET metric. We used sacreCOMET~\cite{zouharchen2024sacrecomet} to create the COMET setup signature.

\section{Results and discussion}
\label{sec:results}
A performance comparison of models on \datasetname~test sets is presented in Table~\ref{table_results}. 

\paragraph{Models with image context}
For the PaliGemma models, those fine-tuned and evaluated with appropriate images outperformed those with unrelated images. Notable improvements were observed in the product names and offer titles sets. However, while the COMET metric showed a decrease for the product descriptions, the chrF metric showed a slight increase.

\paragraph{Models with category context}  
Both fine-tuned models showed improved performance, with the version incorporating category path context achieving a notable advantage in the COMET metric across all datasets. The most significant improvement was observed with product names, while the smallest gain in the COMET metric occurred with product descriptions, where the chrF metric actually decreased.

\paragraph{Models with image descriptions}  
The model using image description prefixes showed a significant decrease in quality, especially on the product description dataset. It is important to note that the fine-tuning was performed on synthetic image descriptions and used a smaller dataset than the models incorporating category context. In this case, the added context had a negative impact on performance, highlighting that fine-tuning an NMT model with prefixes can degrade its quality in some scenarios.

\section{Conclusion}
This study explores methods for incorporating context into MT using the \datasetname~dataset. We are making this dataset publicly available to support research into context-aware translation tasks. We investigated the fine-tuning of VLM for machine translation to exploit image-based context for improved translation quality. Secondly, we analysed the effect of product category paths on translation performance in text-to-text models for e-commerce data. Both experiments showed that the models benefited from contextual information. We also report negative results from fine-tuning with image description prefixes, highlighting that added context can sometimes impair model quality and that this straightforward approach requires further refinement.

\section*{Limitations}

Our approaches rely heavily on the quality of training data and the suitability of the test set for context-aware translation. In many cases, the text alone is sufficient without additional context. Moreover, incorporating extra context can sometimes reduce translation quality, especially with LLMs, where hallucinations may introduce critical errors for users. The experiments with VLM discussed in this paper require significantly more computational resources than text-to-text NMT models, due to the larger model and data sizes. 
We used AI assistance exclusively to enhance the text style and identify grammatical errors in this manuscript.

\bibliography{anthology,custom}

\appendix

\section{Details on experiments}

We conducted all our experiments on a single server equipped with four Nvidia A100 GPUs, each with 80~GB of RAM.

\subsection{VLM setup}
\label{app:vlm_ft}

PaliGemma was fine-tuned using LoRA~\cite{hu2021loralowrankadaptationlarge} with rank $r=8$ and alpha $\alpha=8$. The fine-tuning was performed on a single A100 GPU for 4 epochs with a learning rate of $1e^{-4}$ and batch size set to 16.

\subsection{Text-to-text baseline model}
\label{marian_baseline}
The model was trained on four NVIDIA A100 GPUs. It employs a shared vocabulary of 32,000 subword tokens, generated using the SentencePiece toolkit~\cite{kudo-richardson-2018-sentencepiece}, with all embeddings tied during training. Early stopping was set to 10, with the validation frequency set to 3000 steps and based on the chrF metric on the \datasetname~validation set.

\subsection{Models with category context}
\label{app:category_context}

For fine-tuning with prefixes we employed the following special tokens to mark category context in the SentencePiece vocabulary: 
\begin{itemize} 
    \item <SC> -- start of the category path 
    \item <SEP> -- separator of subcategories 
    \item <EC> -- end of the category path 
    \end{itemize} 
The subcategories were provided in the target language and were not included as special tokens in the vocabulary. An example of a source sentence is as follows: \textit{<SC> Moda <SEP> Odzież, Obuwie, Dodatki <SEP> Obuwie <SEP> Męskie <SEP> Sportowe <EC> Big Star pánské sportovní boty JJ174278 černé 44}. The target sentences remained unchanged.

The model without category context was trained using the same configuration and data setup, but without the category path prefixes. Both fine-tunings were performed on two NVIDIA A100 GPUs with a learning rate of $5e^{-6}$, and early stopping was applied based on the chrF metric on the concatenated \datasetname~validation set.

\subsection{Models with image descriptions}
\label{app:gen_desc}
The image descriptions were generated in the Czech and Polish languages using the~\texttt{google/paligemma-3b-mix-224} model and the \texttt{transformers} library. However, only image descriptions in Czech were used in the experiments. The prompts were structured as simple tasks as found in the original PaliGemma paper~\cite{beyer2024paligemmaversatile3bvlm}. The exact prompts are shown in Table~\ref{tab:img_desc_prompt}. Images for inference were resized to 224x224 pixels.
We include generated Czech and Polish captions in the dataset.
For fine-tuning with image description we employed the following special tokens in the SentencePiece vocabulary: 
\begin{itemize} 
    \item <SD> -- start of the image description
    \item <ED> -- end of the image description
\end{itemize} 
An example of a source sentence is as follows: \textit{<SD> Černé a bílé boty s nápisem " big star " na boku. <ED> Big Star pánské sportovní boty JJ174278 černé 44}

The training configuration was identical to the experiments with category context, except that the validation frequency was reduced to every 300 steps.

\begin{table}[h]
\resizebox{\columnwidth}{!}{%
\begin{tabular}{@{}l|l@{}}
\toprule
Target language & Prompt used for image description \\ \midrule
Czech              & \texttt{popsat obrázek v češtině}         \\
Polish              & \texttt{opisz obrazek po polsku}      \\
\bottomrule
\end{tabular}
}
\caption{Prompts for image description generation used with the paligemma-3b-mix-224 model.}
\label{tab:img_desc_prompt}
\end{table}

\end{document}